\documentclass[10pt,twocolumn,letterpaper]{article}

\usepackage{cvpr}              

\usepackage{multirow}
\usepackage{mathabx}
\usepackage{booktabs}
\usepackage{arydshln}
\usepackage{pifont}
\usepackage{graphicx}
\usepackage[table,xcdraw]{xcolor}
\newcommand{\cmark}{\ding{51}}
\newcommand{\xmark}{\ding{55}}

\usepackage[linesnumbered,ruled,vlined]{algorithm2e}
\usepackage[accsupp]{axessibility}

\usepackage{array}
\usepackage{xparse}
\usepackage{longtable}

\definecolor{cvprblue}{rgb}{0.21,0.49,0.74}
\usepackage[pagebackref,breaklinks,colorlinks,citecolor=cvprblue]{hyperref}

\def\paperID{4033} 
\def\confName{CVPR}
\def\confYear{2024}

\title{MaxQ: Multi-Axis Query for N:M Sparsity Network}

\author{
Jingyang Xiang$^1$
\and
Siqi Li$^1$
\and
Junhao Chen$^1$
\and
Zhuangzhi Chen$^2$
\and
Tianxin Huang$^1$
\and
Linpeng Peng$^1$
\and
Yong Liu$^1$\thanks{Corresponding author}
\and
\centerline{$^1$APRIL Lab, Zhejiang University, Hangzhou, China}
\and
\centerline{$^2$IVSN, Zhejiang University of Technology, Hangzhou, China}
}

\begin{document}
\maketitle
\begin{abstract}

N:M sparsity has received increasing attention due to its remarkable performance and latency trade-off compared with structured and unstructured sparsity.
However, existing N:M sparsity methods do not differentiate the relative importance of weights among blocks and leave important weights underappreciated.
Besides, they directly apply N:M sparsity to the whole network, which will cause severe information loss.
Thus, they are still sub-optimal.
In this paper, we propose an efficient and effective Multi-Axis Query methodology, dubbed as MaxQ, to rectify these problems.
During the training, MaxQ employs a dynamic approach to generate soft N:M masks, considering the weight importance across multiple axes.
This method enhances the weights with more importance and ensures more effective updates.
Meanwhile, a sparsity strategy that gradually increases the percentage of N:M weight blocks is applied, which allows the network to heal from the pruning-induced damage progressively.
During the runtime, the N:M soft masks can be precomputed as constants and folded into weights without causing any distortion to the sparse pattern and incurring additional computational overhead.
Comprehensive experiments demonstrate that MaxQ achieves consistent improvements across diverse CNN architectures in various computer vision tasks, including image classification, object detection and instance segmentation.
For ResNet50 with 1:16 sparse pattern, MaxQ can achieve 74.6\%  top-1 accuracy on ImageNet and improve by over 2.8\% over the state-of-the-art.
Codes and checkpoints are available at \url{https://github.com/JingyangXiang/MaxQ}.
\end{abstract}
\section{Introduction}
\label{sec:intro}

Deep convolutional neural networks~(CNNs) have achieved great success in various computer vision tasks, 
including image classification~\cite{simonyan2015very,he2016deep}, object detection~\cite{he2017mask, girshick2015fast}, semantic segmentation~\cite{girshick2014rich, croitoru2019unsupervised}.
However, the expensive memory and computational overhead have presented challenges for deploying them on mobile or edge devices.
Therefore, it is vital to study the network compression to reduce its runtime overhead while maximally retaining its performance.

\begin{figure}[t]
\centering
\includegraphics[width=0.85\linewidth]{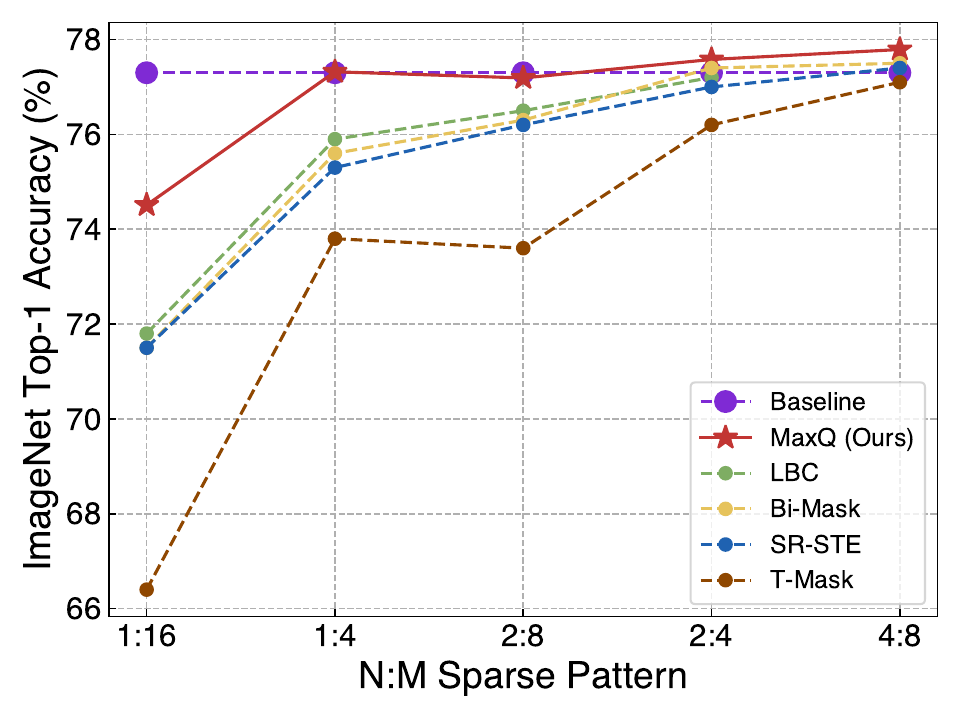}
\caption{Comparison of the accuracy-sparse pattern Pareto curve of the ResNet50 on ImageNet.
MaxQ shows the top-performing Pareto frontier compared with previous N:M sparsity methods~\cite{hubara2021accelerated, zhou2021srste, zhang2022learning, Zhang2023Bimask}.}
\label{fig:pareto}
\vspace{-0.1in}
\end{figure}

%
Among the many compression methods~\cite{hinton2015distilling, han2015deep, howard2017mobilenets, liu2018darts, nagel2021white}, network sparsity stands out as a highly effective tool for achieving practical memory and FLOPs saving.
Most researchers have paid their attention to structured sparsity~\cite{he2018soft, he2019filter, lin2020hrank} and unstructured sparsity~\cite{han2015learning, lee2018snip}.
However, both of them have certain shortcomings in terms of performance and acceleration, limiting their application.
How to realize a better trade-off remains an open problem, as it hinges on factors like the granularity of sparsity and hardware design.

\begin{figure*}[t]
\centering
\includegraphics[width=0.98\linewidth]{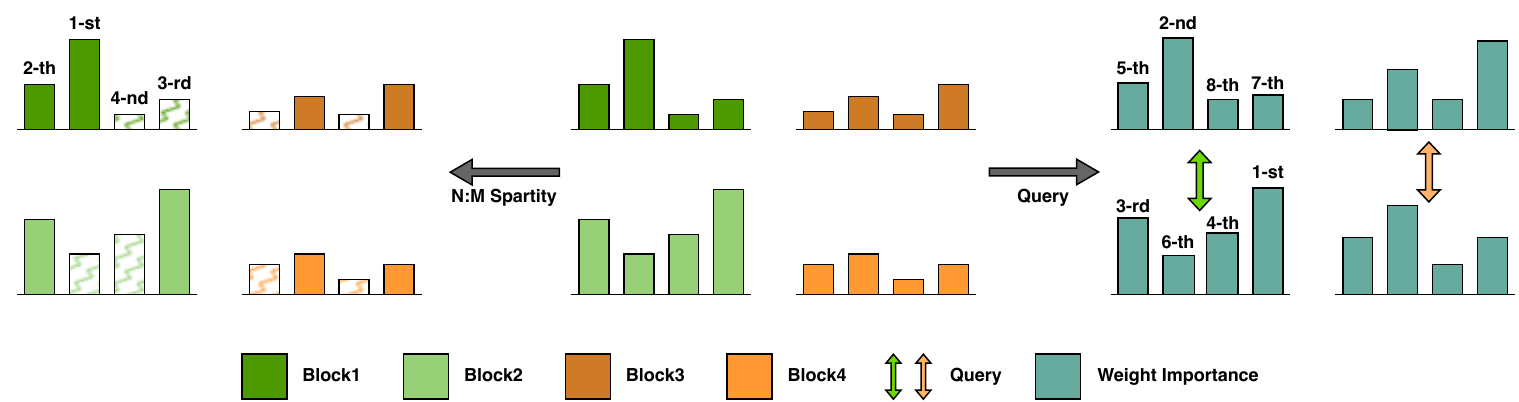}
\caption{The framework of our MaxQ method, which queries the weights importance among the blocks and generates soft masks by querying the weight across multiple axes. For simplify, we only show single axis query.}
\label{fig:distribution}
\vspace{-0.1in}
\end{figure*}

%
Recently, N:M sparsity~\cite{mishra2021accelerating}, a fine-grained sparsity pattern, is considered a highly promising solution.
It can achieve a better trade-off between performance and latency compared to structured and unstructured sparsity thanks to its hardware and software co-design~\cite{mishra2021accelerating, lin2023efficient}.
Several studies have enhanced the performance of N:M sparsity.
The pioneer work ASP~\cite{mishra2021accelerating} proposes to remove the two weights with the smallest magnitude from every four consecutive weights
and use the pretrain-prune-finetune~(PPF) pipeline to get the sparse network.
This approach effectively preserves the performance of the dense network while it still suffers from expensive computation.
To address this, Zhou \etal~\cite{zhou2021srste} proposes to learn N:M sparsity from scratch by extending a regularization term to Straight-Through Estimator~\cite{bengio2013estimating}, which improves the efficiency of sparse architecture update.
Zhang \etal~\cite{zhang2022learning} treats the N:M sparsity problem as a combinatorial problem and solves it in an efficient divide-and-conquer manner.
Although these methods improve the training efficiency or performance of N:M sparsity network, 
they are typically constructed upon the premise of weight block independence and do not differentiate the weight importance among blocks.
At the same time, these methods apply N:M sparsity to the whole network N:M from the beginning of training, leading to critical information loss.
As a result, they are still sub-optimal.

%
In this paper, we propose a simple yet effective \emph{Multi-Axis Query} methodology, dubbed as \textbf{MaxQ}, 
to identify the weight importance and build high-performance N:M sparsity network.
In contrast to the previous approaches, which considered the weights in N:M blocks individually, 
MaxQ queries the weight importance along multiple axes to identify more critical connections and ensures more effective updates for them.
It is worth saying MaxQ achieves this according to the threshold from weight magnitude via a dynamic and parameter-free approach, which is different from the previous methods achieving this by learning a threshold parameter.
Therefore, it requires no additional learnable parameters and simplifies the training process.
Additionally, MaxQ follows an incremental pruning schedule, which gradually increases the percentage of N:M sparse block based on the epoch.
It progressively allows the network to heal from the pruning-induced damage and notably enhances performance.
This strategy makes the training process more stable and weights to be trained more sufficiently, thus improving convergence and performance.
With only one training pass from scratch, our obtained sub-models perform better than the previous method across diverse CNN architectures in various computer vision tasks.
Moreover, our MaxQ can achieve good results in Post-Training-Quantization~(PTQ) even though it is a self-structured re-parameterized network, contrary to the previous structured re-parameterized network.

The contributions of our work are highlighted as follows:

\begin{itemize}
    \item We propose a multi-axis query method MaxQ with two novel features: 
    (1) a multi-axis query approach to identify important connections among the N:M sparse blocks;
    (2) a dynamic approach to generate soft pruning masks in a parameter-free manner.
    \item Our method follows an incremental pruning schedule by increasing the percentage of N:M blocks according to their $\ell_1$-norm.
    It enhances the performance of the N:M sparsity network in one training pass.
    \item Experimentally, MaxQ achieves consistent improvement across different N:M sparse patterns compared to previous methods~(as shown in \cref{fig:pareto}) and is applicable to downstream tasks for computer vision. 
    For image classification on ImageNet, MaxQ achieves 74.6\% top-1 accuracy for a 1:16 sparse ResNet50, improving the previous best~\cite{zhang2022learning} by 2.8\%.
    For object detection and instance segmentation on the COCO~\cite{lin2014microsoft} dataset, MaxQ can achieve comparable results with a dense baseline model under the 1:4 structured sparsity.
    \item MaxQ is friendly to quantization. The ResNet50 with 2:4 sparsity achieves 0.5\% drop~(77.6\%$\rightarrow$77.1\%) in the top-1 accuracy when quantized to INT8 using PTQ, which benefits its deployment.
\end{itemize}

\section{Related Work}
\label{sec:related_work}

\subsection{Sparsity Granularity in Network Compression}

\begin{table}[ht]
\centering
\scriptsize
\begin{tabular}{@{}lllcc@{}}
\toprule
Granularity & Performance                              & Latency                & Hardware                  & Library       \\ 
\midrule
Filter      & \cmark                                   & \cmark \cmark \cmark   & General                   & General       \\
Weight      & \cmark \cmark \cmark                     & \cmark                 & Specific                  & Specific      \\
1$\times$N  & \cmark \cmark                            & \cmark \cmark          & General                   & Specific      \\
N:M         & \cmark \cmark                            & \cmark \cmark \cmark   & Ampere GPUs               & General       \\ \bottomrule
\end{tabular}
\caption{An overview of mainstream sparsity granularity.}
\label{tab:sparse_granularity}
\vspace{-0.2in}
\end{table}

Sparsity is an essential tool for network compression and has attracted extensive attention from both industry and academia.
According to the granularity of sparse weights, network sparsity can be categorized into structured sparsity~\cite{wen2016learning, he2017channel, lin2020hrank}, unstructured sparsity~\cite{han2015learning, guo2016dynamic, frankle2018the, lee2018snip, evci2020rigging, kusupati2020soft} and semi-structured sparsity~\cite{elsen2020fast, mishra2021accelerating, lin20221xn, chenrgp2023}.
\cref{tab:sparse_granularity} overviews these sparsity and their characteristics.
Structured sparsity removes the entire channel of a convolutional or fully connected layer, which enables significant speedup on general-purpose hardware.
Unstructured sparsity removes single weight and achieves negligible performance loss even under a large sparse ratio~\cite{evci2020rigging, kusupati2020soft}.
However, the former suffers huge performance degradation at high compression ratios, and the latter gains rare speedup on general-purpose hardware for its expensive memory access and low computational density resulting from irregular sparse tensors.
In recent years, semi-structured sparsity has received extensive attention from researchers.
By limiting the distribution of weights to specific sparse types, semi-structured sparsity can achieve a better trade-off between performance and latency through a synergistic software and hardware design.
Among the semi-structured sparsity, 1$\times$N~\cite{elsen2020fast, lin20221xn} and N:M~\cite{mishra2021accelerating} are now widely supported by software and hardware.
1$\times$N makes N consecutive output channels keep the same sparse pattern, enabling significant speedup on general-purpose hardware.
N:M sparsity forces at most N of the M consecutive weights along the input channel dimension to be non-zero, achieving good results while maintaining high sparsity and high-performance acceleration on NVIDIA Ampere GPUs.

The core purpose of network sparsity is to remove unimportant weights from the network and preserve the network's performance.
Therefore, learning the appropriate mask is indeed necessary.
Mask is usually obtained from the weight values according to specific criteria, and the process must follow a schedule.
We empirically categorize existing studies into two groups below based on criteria and schedule.

\noindent \textbf{Mask Criteria.}
Extensive studies have investigated how to identify the critical weights in a network.
Among them, the most widely used is the magnitude-based criterion that selects the pruning targets by their absolute value or $\ell_1$-norm.
Apart from this, He \etal~\cite{he2019filter} proposed to prune filters via geometric median to compress CNN models with redundancy rather than those with "relatively less" importance.
Molchanov \etal~\cite{molchanov2019importance} proposed to estimate the contribution of a neuron (filter) to the final loss via the first and second-order Taylor expansions.
Lee \etal~\cite{lee2018snip} introduced a saliency criterion based on connection sensitivity to identify structurally important connections in the network.
Some other criteria, such as BN-based~\cite{liu2017learning}, have also proven effective.

\noindent \textbf{Mask Schedule.}
Learning schedule for masks is also vital for a sparse network.
Most of the early research~\cite{han2015learning,lin2020hrank} follows the PPF pipeline iteratively, which is complicated to use and time-consuming.
In order to simplify this process, recent studies use soft and incremental pruning to train sparse networks from scratch without dependence on pretrained weights.
For example, Humble \etal~\cite{humble2022soft} proposed soft masking for channel pruning, which allows pruned channels to adaptively return to the network while simultaneously pruning towards a target cost constraint.
Hou \etal~\cite{hou2022chex} proposed a channel exploration methodology to prune and regrow the channels with a sampling-based strategy repeatedly.
Evci~\cite{evci2020rigging} introduced a sparse-to-sparse training procedure with a fixed parameter count and a fixed computational cost, sacrificing no accuracy relative to existing dense-to-sparse training methods.
Compared to hard and one-shot pruning, these soft and incremental strategies can improve the model's performance, especially at a high sparse ratio.

\section{Methodology}
\label{sec::methodology}
\subsection{Preliminaries}

Here, we define the N:M sparsity problem. Without loss of generality,
we suppose an L-layer CNN with parameters $\mathbf{W}=\{\mathbf{w}^1,...,\mathbf{w}^L\}$, with $\mathbf{w}^l \in \mathbb{R}^{C^l_{\text{out}}\times C_{\text{in}}^l\times K^l_{\text{h}} \times K^{l}_{\text{w}}}$. $C^l_{\text{out}}$, $C^l_{\text{in}}$, $K^l_{\text{h}}$ and $K^l_{\text{w}}$ represent the number of output channels, input channels, kernel height and kernel width for $l$-th layer respectively. 
N:M sparsity groups every M consecutive weights along the input channel in each layer and requires each group to have at most N non-zero elements.
With this pattern, only N non-zero values in each group need to be stored and metadata is adopted to encode the position of each non-zero value.
This process can compress the origin matrix and be accelerated by designated processing units~(\eg NVIDIA Ampere Tensor Cores).
To this end, we first rearrange $\mathbf{w}^l$ to ${\mathbf{m}^l\in \mathbb{R}^{G^l\times M}}$, where $G^l=\frac{C^l_{\text{out}}\times K^l_{\text{h}} \times K^l_{\text{w}} \times C^l_{\text{in}}}{M}$ and denote $\mathbf{M}=\{\mathbf{m}^1,...,\mathbf{m}^L\}$.
Meanwhile, N:M binary masks $\mathbf{b}^l\in \left \{ 0,1 \right \} ^{G^l\times M}$ are introduced to achieve such sparsity
Then, the optimization objective can be formulated as:
\begin{equation}
\label{eq:object}
\min_{\mathbf{M},\mathbf{B}}\mathcal{L}(\mathbf{M}\cdot \mathbf{B};\mathcal{D})~~s.t.  \| \mathbf{b}^l_{g^l,:}  \|_0 \leq N
\end{equation}
where $g^l=1,2,...,G^l$. The $\mathcal{L}\left ( \cdot \right )$ represents the loss function and $\mathcal{D}$ denotes the observed data.

\subsection{Multi-Axis Query}

\begin{figure}[t]
\centering
\includegraphics[width=0.65\linewidth]{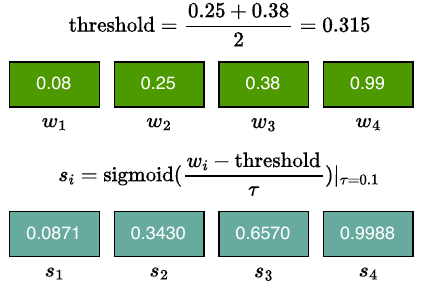}
\vspace{-0.1in}
\caption{The process of computing soft masks. The weights of the model are sorted in descending order based on their magnitudes for clear understanding. We assume $(N, p, \tau)=(4,0.5,0.1)$.}
\label{fig:threshold}
\vspace{-0.1in}
\end{figure}

Although extensive studies have promoted the performance of N:M sparsity, they only focus on picking the N most important weights from M elements and applying binary masks to them, which makes a portion of the weights with more importance underappreciated.
Therefore, it is vital to exploit the importance of weights further to improve the performance of N:M sparsity networks.

To represent the importance of the weights in the N:M sparse pattern, we first relax the constrain of the binary matrix $\mathbf{b}^l$ to $\mathbf{s}^l\in \{x|x\ge 0\}^{G^l\times M}$.
Then, the optimization objective can be rewritten as:
\begin{equation}
\min_{\mathbf{M},\mathbf{S}}\mathcal{L}(\mathbf{M}\cdot \mathbf{S};\mathcal{D})~~s.t.  \| \mathbf{s}^l_{g^l,:} \|_0 \leq N
\label{eq:rewritten}
\end{equation}

It can be seen as replacing the previous hard mask $\mathbf{b}^l$ with a soft mask $\mathbf{s}^l$.
$\mathbf{s}^l$ satisfies the N:M sparse pattern and can be folded into the network as constants, which will not cause any distortion to the sparse pattern and introduce any extra cost during the runtime.

To get the value of $\mathbf{s}^l$, we propose MaxQ to measure the importance of weights.
As shown in \cref{fig:distribution}, MaxQ can be divided into two parts in what follows.

\noindent   
\textbf{Part 1: Apply N:M Sparsity.}

First, for $l$-th layer, we initialize all elements in $\textbf{b}^l$ to 1. For $t$-th epoch, we sort each element in $\mathbf{m}^l_g$ according to it absolute value and sort the $\mathbf{m}^l$ according to it's $\ell_1$-norm  to identify the set of blocks to apply N:M sparsity:
\begin{equation}
\begin{split}
    \mathcal{M}^l_g & = \text{ArgTopK}_{\text{M-N}}\left(-\left|\mathbf{m}_{g,:}^l\right| \right) \\
    \mathcal{T}_t^l & = \text{ArgTopK}_{\left \lceil G^{l}\delta_t \right \rceil}\left(\left \{ \left \| \mathbf{m}^l_{g} \right \|_1 \right \}\right)
\end{split}
\label{eq:get_indices}
\end{equation}
where $\mathcal{M}^l \in \mathbb{R}^{G^l\times (M-N)}$,
$\mathcal{T}_k^l \in \mathbb{R}^{\left \lceil G^{l}\delta_t \right \rceil}$
and $\delta_t$ represents the percentage of weight blocks to apply N:M sparsity, which gives the indices of weights with the $(M-N)$ smallest absolute value in each block and blocks with the top $\left \lceil G^{l}\delta_t \right \rceil$ value in $\ell_1$-norm respectively. 
Then we prune the weight by zeroizing $\left\{\textbf{b}^l_{i,j} | i\in\mathcal{T}^l_{t},j\in \mathcal{M}^{l}_{i}\right\}$.

\noindent
\textbf{Part 2: Measure Importance.}
We show our function $S=G(V,p,\tau)$ to measure weight importance. 
Give a vector $V\in \mathbb{R}^N$ and a sparse rate $p$, we get the threshold $\sigma$ by
\begin{equation}
\left\{\begin{split}
\sigma_h&=\text{min}\left(\text{topK}\left(\text{abs}(V),(1-p)\cdot N\right)\right) \\
\sigma_l~&=\text{max}\left(\text{topK}\left(-\text{abs}(V),p\cdot N\right)\right) \\
\sigma~~&=\left(\text{abs}(\sigma_h)+\text{abs}(\sigma_l)\right )/2
\end{split}\right.
\label{eq:compute_threshold}
\end{equation}

Then, we measure weight importance with sigmoid:
\begin{equation}
    s_i=\text{sigmoid}\left(\left({|v_i|-\sigma}\right)/{\tau}\right)
    \label{eq:compute_value}
\end{equation}
where $i=1,2,...,N$ and $\tau$ is the global temperature parameter to control the level of softness in the masks.
\cref{fig:threshold} shows the process when $(N, p, \tau)=(4,0.5,0.1)$.

To measure the weight importance among the blocks, we query the weight along filter axis and kernel axis to generate corresponding soft masks:
\begin{equation}
\begin{split}
\textbf{s}_{i,:,:,:}^{l^{(f)}} & = \left (G\left (w^l_{i,:,:,:},\left(M-N\right)/{M},\tau\right)\right ) \\
\textbf{s}_{:,:,k_1,k_2}^{l^{(k)}} & = \left (G\left (w^l_{:,:,k_1,k_2},\left(M-N\right)/{M},\tau\right)\right )
\end{split}
\label{eq:compute_soft_masks}
\end{equation}
where $i=1,2,...,C_{\text{out}}^l$, $k_1=1,...,K^l_h$ and $k_2=1,...,K^l_w$.

\begin{figure}[t]
\centering
\includegraphics[width=0.90\linewidth]{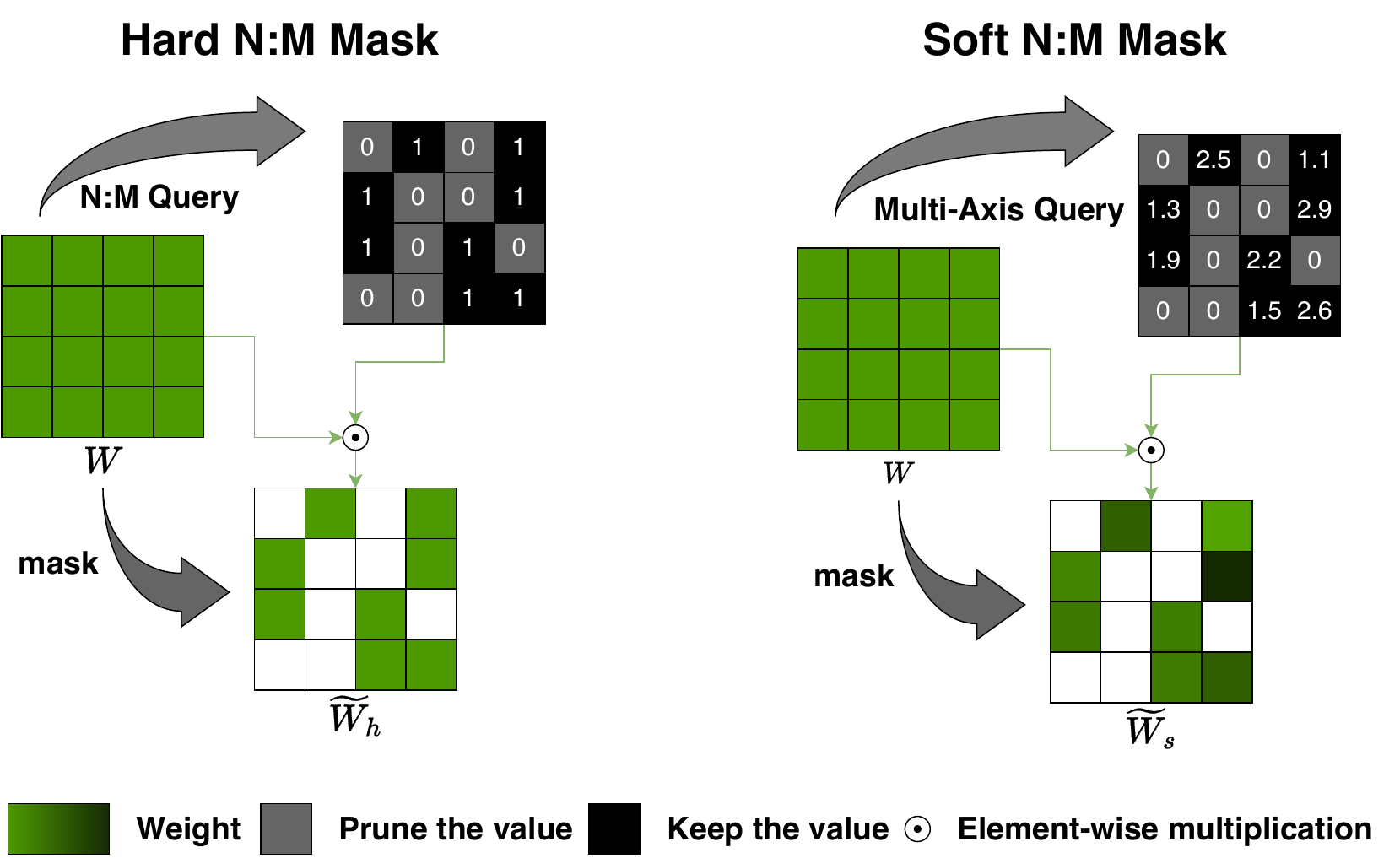}
\vspace{-0.05in}
\caption{MaxQ to generate soft masks.}
\label{fig:soft_mask}
\vspace{-0.2in}
\end{figure}

We rearrange~(RA) the $\textbf{s}^{l^{(f)}}$ and $\textbf{s}^{l^{(k)}}$ to match the shape of $\mathbf{b}^l$ 
and obtain the soft mask as shown in \cref{fig:soft_mask}:
\begin{equation}
\begin{split}
    \mathbf{s}^l & = \mathbf{b}^l+\mathbf{b}^l\odot\text{RA}(\textbf{s}^{l^{(f)}})+\mathbf{b}^l\odot\text{RA}(\textbf{s}^{l^{(k)}}) \\
    & = \mathbf{b}^l \odot \left (1+\text{RA}(\textbf{s}^{l^{(f)}})+\text{RA}(\textbf{s}^{l^{(k)}})\right)
\end{split}
\label{eq:get_final_mask}
\end{equation}
It is worth noting $\textbf{s}^l$ can be precomputed.
Meanwhile, since $\textbf{b}^l$ satisfies the N:M sparse pattern, $\textbf{s}^l$ will also satisfy this pattern.
Therefore, MaxQ will not bring any additional overhead during the runtime compared to the conventional N:M sparsity network.

\subsection{Incremental Sparsity}
Previous methods fixed $\delta_t$ to 1 throughout the training process, which means the whole network is in N:M sparse pattern from the beginning of training.
It will cause severe information loss and is detrimental to network convergence.

To this end, incremental sparsity, which gradually increases the sparse ratio based on the current epoch/step, has been shown to be an effective technique to heal sparse networks from pruning and improve their performance.
For unstructured sparsity, incremental sparsity can be achieved by gradually removing a single weight; for structured sparsity, it can be achieved by removing the channels across layers.
Since networks often have a large number of parameters and channels, the sparsity process can be viewed as continuous.
Different from them, N:M sparse is characterized by several M consecutive weight groups.
If we increase the sparsity by simply reducing $\left \| \mathbf{b}^l_{g,:} \right \|_0$ from $M$ to $N$, the sparse networks will suffer critical performance degradation, especially at high sparsity.

In this paper, we propose gradually increasing the percentage of N:M sparse blocks to achieve a smoother sparsity process.
We apply the same sparse schedule for each layer
The percentage of N:M sparse blocks at the $t$-th training epoch is computed as:
%
\begin{equation}
    \label{eq:incre_sparse}
    \delta_t=\min(1, \max(0, 1-\left[1-(t-t_i)/(t_f-t_i)\right]^3)) 
\end{equation}
where $t_i$ and $t_f$ denote the beginning and ending epochs in the incremental sparsity process.
Specifically, if $t$ is smaller than $t_i$, the network is trained in a dense state, and when $t$ is larger than $t_f$, the network is trained in a N:M sparse pattern.
Notably, we firstly apply the N:M sparse pattern for blocks with larger $\ell_1$-norm because we find 
it can reduce the performance degradation when the sparsity increases and keep the convergence process stable.
We conduct ablation studies in \cref{ablation} to show its advantage.
As an algorithm guideline, the pseudo-code of MaxQ is provided in \cref{algorithm:overview}.
%
%
%

\setlength{\textfloatsep}{0.1cm}
\begin{algorithm}[t]
\small
    \textbf{Input}: An $L$-layer CNN model with weights $\mathbf{W}=\{\mathbf{w}^1,...,\mathbf{w}^L\}$; target sparse pattern $\text{N}$:$\text{M}$; total training epochs $T_{\text{total}}$; initial and final epoch for incremental pruning $t_i$, $t_f$; training set $\mathcal{D}$ \;
    \textbf{Output}: A sub-model satisfying the target sparse pattern $\text{N}$, $\text{M}$ and its optimal weight values $\mathbf{W}^*$\;
    Randomly initialize the model weights $\mathbf{W}$\;
    Rearrange the model weights $\mathbf{W}$ to $\mathbf{M}$ \;
    \For{each training epoch $t\in[1, ..., T_\text{total}]$}
    {
        Compute the percentage of N:M blocks $\delta_t$ via \cref{eq:incre_sparse}\;
        \For{each mini-batch $\in$ $\mathcal{D}$}{
            \For{$l \in [1,..., L]$}{
                Reset $\left\{\mathbf{b}_{i,j}^{l} | \forall i,\forall j\right\}$ to 1 \;
                Get the indices $\mathcal{M}^l$ and $\mathcal{T}_t^l$ via \cref{eq:get_indices} \;
                Set $\left\{\textbf{b}^l_{i,j} | i\in\mathcal{T}^l_{t},j\in \mathcal{M}^{l}_{i}\right\}$ to 0 \;
                Get the $\textbf{s}^{l^{(f)}}$ and $\textbf{s}^{l^{(k)}}$ via \cref{eq:compute_threshold} $\sim$ \cref{eq:compute_soft_masks} \;
                Get the soft mask $\mathbf{s}^l$ via \cref{eq:get_final_mask}
            }
            Forward via \cref{eq:rewritten} \;
            Backward and update via the SGD optimizer \;
        }
    }
    Compute $\mathbf{M}^*=\{\textbf{m}^l \cdot \textbf{s}^l | ~\forall l~ \}$ \;
    Rearrange $\mathbf{M}^*$ back to $\mathbf{W}^*$ \;
\caption{Overview of the MaxQ method.}
\label{algorithm:overview}
\end{algorithm}
\setlength{\floatsep}{0.1cm}
\section{Experiment}
\label{sec::experiment}
\subsection{Experiment Settings}

\begin{table}[htbp]
\scriptsize
\centering
{
\begin{tabular}{@{}llccccc}
\toprule
Model                               & Method        & N:M         & Top-1           & Epochs & FLOPs    & Params \\ \midrule
\multirow{9}{*}{ResNet34~~~~~~~}    & Baseline      & -           & 74.6\%          & 120    & 3.67G     & 21.8M      \\ \cmidrule(l){2-7} 
                                    & ASP           & 1:4         & 70.9\%          & 200    & 1.01G     & 5.85M      \\
                                    & SR-STE        & 1:4         & 73.8\%          & 120    & 1.01G     & 5.85M      \\
                                    & LBC           & 1:4         & 73.7\%          & 120    & 1.01G     & 5.85M      \\ 
                                    & \bf MaxQ          & 1:4         & \bf 74.2\%          & 120    & 1.01G     & 5.85M      \\ \cmidrule(l){2-7}                                  
                                    & ASP           & 2:4         & 73.9\%          & 200    & 1.90G     & 11.2M      \\
                                    & SR-STE        & 2:4         & 74.3\%          & 120    & 1.90G     & 11.2M      \\
                                    & LBC           & 2:4         & 74.1\%          & 120    & 1.90G     & 11.2M      \\ 
                                    & \bf MaxQ          & 2:4         & \bf 74.5\%          & 120    & 1.90G     & 11.2M      \\ \midrule
\multirow{17}{*}{ResNet50~~~~~~~}   & Baseline      & -           & 77.3\%          & 120    & 4.11G     & 25.6M      \\ \cmidrule(l){2-7} 
                                    & ASP           & 2:4         & 77.4\%          & 200    & 2.12G     & 13.8M      \\
                                    & SR-STE        & 2:4         & 77.0\%          & 120    & 2.12G     & 13.8M      \\
                                    & LBC           & 2:4         & 77.2\%          & 120    & 2.12G     & 13.8M      \\ 
                                    & \bf MaxQ          & 2:4         & \bf 77.6\%          & 120    & 2.12G     & 13.8M      \\ \cmidrule(l){2-7} 
                                    & ASP           & 1:4         & 76.5\%          & 200    & 1.11G     & 7.93M      \\
                                    & SR-STE        & 1:4         & 75.3\%          & 120    & 1.11G     & 7.93M      \\
                                    & LBC           & 1:4         & 75.9\%          & 120    & 1.11G     & 7.93M      \\ 
                                    & \bf MaxQ          & 1:4         & \bf 77.3\%          & 120    & 1.11G     & 7.93M      \\ \cmidrule(l){2-7}
                                    & ASP           & 2:8         & 76.6\%          & 200    & 1.11G     & 7.93M      \\
                                    & SR-STE        & 2:8         & 76.2\%          & 120    & 1.11G     & 7.93M      \\
                                    & LBC           & 2:8         & 76.5\%          & 120    & 1.11G     & 7.93M      \\ 
                                    & \bf MaxQ          & 2:8         & \bf 77.2\%          & 120    & 1.11G     & 7.93M      \\ \cmidrule(l){2-7} 
                                    & ASP           & 1:16        & 71.5\%          & 200    & 0.44G     & 3.52M      \\
                                    & SR-STE        & 1:16        & 71.5\%          & 120    & 0.44G     & 3.52M      \\
                                    & LBC           & 1:16        & 71.8\%          & 120    & 0.44G     & 3.52M      \\ 
                                    & \bf MaxQ          & 1:16        & \bf 74.6\%          & 120    & 0.44G     & 3.52M      \\
\bottomrule
\end{tabular}
}
\caption{Results of the different N:M sparsity training methods for ResNet34 and ResNet50 on ImageNet.}
\label{tab:resnet}
\end{table}

\begin{table}[ht]
\scriptsize
\centering
{
\begin{tabular}{@{}llccccc}
\toprule
Model                        & Method   & N:M         & Top-1           & Epochs & FLOPs     & Params    \\ \midrule
\multirow{7}{*}{MobileNetV1} & Baseline & -           & 71.9\%          & 120    & 578M      & 4.23M      \\ \cmidrule(l){2-7} 
                             & ASP      & 2:4         & 70.4\%          & 200    & 302M      & 2.66M      \\
                             & SR-STE   & 2:4         & 71.5\%          & 120    & 302M      & 2.66M      \\ 
                             & \bf MaxQ     & 2:4         & \bf 72.1\%          & 120    & 302M      & 2.66M      \\ \cmidrule(l){2-7} 
                             & ASP      & 1:4         & 65.4\%          & 200    & 164M      & 1.88M      \\
                             & SR-STE   & 1:4         & 67.8\%          & 120    & 164M      & 1.88M      \\ 
                             & \bf MaxQ     & 1:4         & \bf 68.5\%          & 120    & 164M      & 1.88M      \\ \bottomrule
\end{tabular}}
\caption{Results of the different N:M sparsity training methods for lightweight model MobileNetV1 on ImageNet.}
\label{tab:mobile}
\end{table}

To validate the effectiveness of MaxQ, we conducted image classification on ImageNet with heaveweight CNNs ResNet34~\cite{he2016deep}, ResNet50~\cite{he2016deep} and lightweight MobileNetV1~\cite{howard2017mobilenets}.
We compare MaxQ with N:M sparsity and unstructured sparsity in \cref{structured_sparsity} and \cref{unstructured_sparsity}, respectively.
We also conducted object detection and semantic segmentation on the COCO~\cite{lin2014microsoft} benchmark with Faster-RCNN~\cite{ren2015faster} and Mask-RCNN~\cite{he2017mask} in \cref{coco}.
All experiments are implemented on PyTorch with NVIDIA RTX 3090
and trained with the same configurations as previous work~\cite{zhou2021srste, zhang2022learning} to make a fair comparison.
$t_i$ and $t_f$ are set to 0 and 3/4 of the total training epochs respectively.
Ablation studies about components, $t_i$ and $t_f$ and strategy for incremental sparsity are demonstrated in \cref{ablation}.
Performance analysis is shown in \cref{sec:performance}.

%


\subsection{Comparison with N:M sparsity}
\label{structured_sparsity}

We first apply our MaxQ to ResNet34 and ResNet50 to validate its effectiveness.
As shown in \cref{tab:resnet}, MaxQ leads all N:M sparse patterns and networks.
For ResNet34, MaxQ outperforms SR-STE~\cite{zhou2021srste} and LBC~\cite{zhang2022learning} at 1:4 sparse pattern at the
top-1  accuracy by 0.4\% and 0.5\% respectively.
For ResNet50, MaxQ achieves similar improvement at 1:4 and 2:8 sparse patterns.
Meanwhile, we also conduct experiments on lightweight CNN MobileNetV1, as compressing this lightweight network is more beneficial for further acceleration on mobile devices.
Similar to ResNet, we apply N:M sparsity to all except the first, last layers and depthwise convolutional layers.
The results in \cref{tab:mobile} also demonstrate the superiority against the others.

The results in \cref{tab:resnet} and \cref{tab:mobile} also show two interesting properties of our MaxQ method.
ResNet34, ResNet50 and MobileNetV1 achieve 74.5\%, 77.6\% and 72.1\% top-1 accuracy at 2:4 sparse pattern, while their dense counterpart achieves 74.6\%, 77.3\% and 71.9\% respectively, which means our MaxQ can achieve almost lossless or better results on various networks at 2:4 sparse pattern.
On the other hand, our MaxQ achieves a more significant improvement at a high sparse ratio.
For example, our 1:16 ResNet50 achieves 74.6\% top-1 accuracy, which outperforms the previous state-of-the-art by 2.8\%.

\begin{table}[t]
        \scriptsize
	\centering
        {
	\begin{tabular}[b]{lccccccc}
		\toprule
		  Method      & Top-1               & Sparsity        & FLOPs         & Params         & S             & U \\ \midrule
		  Baseline    & 77.3\%              & 0.0             & 4.10G          & 25.6M           & -             & -       \\ \midrule
            RigL~\cite{evci2020rigging}        & 74.6\%              & 80              & 0.92G          & 5.12M           & \xmark        & \cmark \\
            GMP~\cite{zhu2017prune}         & 75.6\%              & 80              & 0.82G          & 5.12M           & \xmark        & \cmark \\
            MAP~\cite{back2023magnitude}         & 75.9\%              & 80              & -              & 5.12M           & \xmark        & \xmark \\ 
            STR~\cite{kusupati2020soft}         & 76.2\%              & 81              & 0.82G          & 5.12M           & \xmark        & \cmark \\ \midrule
            SR-STE      & 75.3\%              & 1:4             & 1.13G          & 7.97M           & \cmark        & \cmark \\ 
            LBC         & 75.9\%              & 1:4             & 1.13G          & 7.97M           & \cmark        & \cmark \\ 
            \bf MaxQ        & \bf 77.3\%              & 1:4             & 1.13G          & 7.97M           & \cmark        & \cmark \\ \midrule
            SR-STE      & 76.2\%              & 2:8             & 1.13G          & 7.97M           & \cmark        & \cmark \\
            LBC         & 76.5\%              & 2:8             & 1.13G          & 7.97M           & \cmark        & \cmark \\ 
            \bf MaxQ        & \bf 77.2\%              & 2:8             & 1.13G          & 7.97M           & \cmark        & \cmark \\ \midrule
		  DNW~\cite{wortsman2019discovering}         & 68.3\%              & 95              & 0.20G          & 1.28M           & \xmark        & \xmark \\
		  RigL~\cite{evci2020rigging}        & 70.0\%              & 95              & 0.49G          & 1.28M           & \xmark        & \cmark \\
		  GMP~\cite{zhu2017prune}         & 70.6\%              & 95              & 0.20G          & 1.28M           & \xmark        & \cmark \\
		  STR~\cite{kusupati2020soft}         & 70.4\%              & 95              & 0.16G          & 1.24M           & \xmark        & \xmark \\ 
            OptG~\cite{zhang2022optg}        & 72.5\%              & 95              & 0.22G          & 1.28M           & \xmark        & \xmark \\ \midrule
            SR-STE      & 71.5\%              & 1:16            & 0.44G          & 3.52M           & \cmark        & \cmark \\
    	LBC         & 71.8\%              & 1:16            & 0.44G          & 3.52M           & \cmark        & \cmark \\  
            \bf MaxQ        & \bf 74.6\%              & 1:16            & 0.44G          & 3.52M           & \cmark        & \cmark \\ \bottomrule
	\end{tabular}}
    \caption{Results of the N:M and unstructured sparsity methods for ResNet50 on ImageNet. S: Structured. U: Uniform.}
    \label{tab:unstructure}
\end{table}

\subsection{Comparison with Unstructured Sparsity}
\label{unstructured_sparsity}
We also compare the performance of MaxQ with state-of-the-art unstructured sparsity methods using ResNet50.
The results in \cref{tab:unstructure} demonstrate that our MaxQ consistently achieves outstanding accuracy under various sparsity constraints.
For example, ResNet50 with 1:4 and 2:8 sparse patterns achieve 77.3\% and 77.2\% top-1 accuracy respectively.
Meanwhile, ResNet50 with 1:16 sparse pattern achieves 74.6\% accuracy, surpassing STR and OptG over 4.2\% and 2.1\%.
In contrast to unstructured sparsity, N:M organizes weights in a structured format, avoiding inefficient memory access and low computational density due to irregular distribution of weights.
Additionally, the uniform distribution of N:M sparse weights ensures that computational loads are balanced when performing parallel computation and avoiding speed degradation due to load imbalance across threads.
These exhibit the effectiveness and advantages of exploring N:M sparsity.

\begin{table}[t]
\scriptsize
\centering
\setlength{\tabcolsep}{4mm}
{
\begin{tabular}{@{}clcc}
\toprule
Model                   & Method   & N:M            & mAP       \\ \midrule
\multirow{7}{*}{F-RCNN} & Baseline & -              & 37.4      \\ \cmidrule(l){2-4}
                        & SR-STE   & 2:4            & 38.2      \\
                        & LBC      & 2:4            & 38.5      \\ 
                        & \bf MaxQ     & 2:4            & \bf 38.7      \\ \cmidrule(l){2-4} 
                        & SR-STE   & 1:4            & 37.2      \\
                        & LBC      & 1:4            & 37.3      \\ 
                        & \bf MaxQ     & 1:4            & \bf 37.7      \\ \bottomrule
\end{tabular}}
\caption{Results for object detection on COCO benchmark.}
\label{tab:object}
\end{table}

\begin{table}[t]
    \vspace{0.1in}
    \scriptsize
    \centering
    {\begin{tabular}{@{}clccc}
    \toprule
    Model                   & Method    & N:M         & Box mAP  & Mask mAP  \\ \midrule
    \multirow{7}{*}{M-RCNN} & Baseline  & -           & 38.2     & 34.7      \\ \cmidrule(l){2-5}
                            & SR-STE    & 2:4         & 39.0     & 35.3      \\
                            & LBC       & 2:4         & \bf 39.3 & 35.4      \\ 
                            & \bf MaxQ  & 2:4         & 39.2     & \bf 35.5  \\ \cmidrule(l){2-5} 
                            & SR-STE    & 1:4         & 37.6     & 33.9      \\
                            & LBC       & 1:4         & 37.8     & 34.0      \\ 
                            & \bf MaxQ  & 1:4         & \bf 38.3 & \bf 34.4  \\ \bottomrule
    \end{tabular}}
    \caption{Results for instance segmentation on COCO benchmark.}
    \label{tab:seg}
\end{table}

\subsection{Object Detection and Instance Segmentation}
\label{coco}
In addition to the image classification, we conducted experiments on the challenging dataset COCO based on MMDetection~\cite{mmdetection}, to exploit the generalization ability of MaxQ.
For object detection, we employ classical models Faster R-CNN, and for instance segmentation, we use Mask R-CNN.
%
%
\cref{tab:object} demonstrates that MaxQ consistently outperforms previous methods in object detection.
For example, MaxQ yields a robust performance of 38.7 mAP at 2:4 sparse pattern,
which exceeds the previous state-of-the-art LBC by 0.2 mAP and improves it's dense counterpart by 1.3 mAP.
Similar trends are observed in instance segmentation, as shown in \cref{tab:seg}. 
Furthermore, MaxQ delivers results comparable to dense baseline models at a 1:4 structure sparsity.
These results suggest that N:M sparse networks exhibit similar or better feature transfer capabilities than dense networks and can be effectively applied in downstream tasks.

\subsection{Ablation Study}
\label{ablation}
\vspace{-0.1in}

\begin{table}[h]
\centering
\scriptsize
\begin{tabular}{@{}lc@{}}
\toprule
Method                                                       &  ~~~~Top-1~~~~ \\ 
\midrule
Baseline~(SR-STE)                                            &   71.5\%        \\
~~~+~Multi-Axis Query                                        &   74.2\%        \\
~~~+~Incremental Pruning~(\textbf{default})~~~~~~~~          &   74.6\%        \\
~~~+~Train 200 Epochs                                        &   75.2\%        \\
\bottomrule
\end{tabular}
\caption{Ablation study of different components in MaxQ.}
\label{tab:compent}
\vspace{-0.1in}
\end{table}

We investigate the effectiveness of different components in the MaxQ through ablation studies.
All the following results are based on the ResNet50 model with 1:16 sparse pattern on the ImageNet dataset.

\noindent
\textbf{Components.}
In \cref{tab:compent}, we investigate the effectiveness of different components in MaxQ, namely multi-axis query and incremental pruning.
The baseline is derived from SR-STE where the model is trained solely with the sparse-refined straight-through estimator.
There is no weight importance identifying among blocks and incremental pruning stage.
The multi-axis query identifies and enhances the weights with more importance using soft masks, preventing crucial weights from being overlooked.
It ensures more effective updates for important weights during training, resulting in 2.7\% accuracy improvement.
When the percentage of N:M sparse block in each layer increases according to a specific schedule, instead of being fixed to 100\%  early in training, we can obtain a more optimal N:M sparse network.
This dynamic approach further improves accuracy by 0.4\%.
Additionally, to achieve a fair comparison with ASP, we conduct an experiment to study the effect of training epochs.
We train MaxQ for 200 epochs, the same ASP setup.
The experiment result shows that longer training epochs lead to considerable performance gains, improving accuracy by 0.6\% compared to the default settings.
It's worth noting that for N:M sparse networks, the PPF training pipeline is unnecessary as it does not significantly improve its performance.
On the contrary, applying soft and incremental strategies to train an N:M sparse network from scratch yields better results.
Moreover, these strategies are more straightforward when compared to the PPF pipeline.

\begin{table}[h]
\centering
\scriptsize
\begin{tabular}{@{}lcccccc@{}}
\toprule
Model                     & N:M  & $t_i$ & $t_f$ & $t_f-t_i$  & Top-1    & FLOPs(Train)\\ \midrule
\multirow{6}{*}{ResNet50} & 1:16 & 0     & 30    & 30         & 74.51\%   & 0.74$\times$(3.2e18)\\
                          & 1:16 & 0     & 60    & 60         & 74.41\%   & 0.78$\times$\\
                          & 1:16 & 0     & 90    & 90         & 74.55\%   & 0.81$\times$\\
                          & 1:16 & 30    & 60    & 30         & 74.56\%   & 0.81$\times$\\
                          & 1:16 & 30    & 90    & 60         & 74.52\%   & 0.85$\times$\\
                          & 1:16 & 60    & 90    & 30         & 74.47\%   & 0.89$\times$\\ \bottomrule
\end{tabular}
\caption{Ablation study of different $t_i$ and $t_f$ in MaxQ.}
\label{tab:t_i_t_f}
\vspace{-0.1in}
\end{table}

\noindent
\textbf{Choice of \bf{$t_i$} and \bf $t_f$.}
We conduct ablation studies involving different $t_i$ and $t_f$ in MaxQ.
To explain, larger $t_i$ and $t_f-t_i$ indicate more weights can be sufficiently trained before applying N:M sparsity and a smoother sparsity process.
%
%
%
%
And the total train FLOPs are positively correlated with $t_i$ and $t_f$.
As illustrated in \cref{tab:t_i_t_f}, our method can achieve excellent performance across various $t_i$ and $t_f$.
Moreover, more training FLOPs does not bring a notable improvement to the final result.
It suggests that the weight distribution characteristics of the trained dense network are not essential and explains why the PPF strategy for the ASP does not yield favorable results in the N:M sparsity networks. 
Superior N:M sparsity networks can be trained in one training pass by soft, dynamic and incremental strategies.


\begin{figure}[t]
\centering
\includegraphics[width=0.8\linewidth]{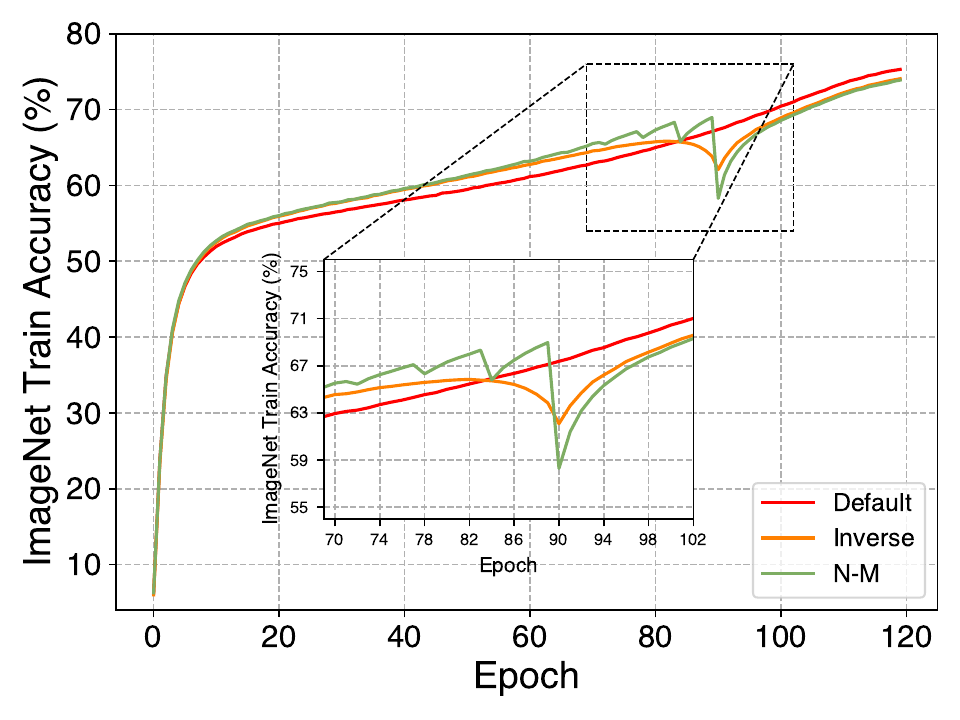}
\vspace{-0.15in}
\caption{Convergence visualization for different strategies. Inverse means we firstly apply the N:M sparsity for blocks with smaller $\ell_1$-norm.
N-M means reducing $\left \| \mathbf{b}^l_{g,:} \right \|_0$ from M to N.}
\label{fig:sort_convergence}
\end{figure}

%
\noindent
\textbf{Sparsity strategy.}
We introduce two other strategies to investigate the efficiency of our strategy for incremental sparsity.
$t_i$ and $t_f$ are set to 0 and 90 respectively.
As shown in \cref{fig:sort_convergence}, both of inverse and N-M suffer severe performance degradation when the sparse ratio increases~(when epoch increases to 90).
Applying N:M sparsity first to weight blocks with larger $\ell_1$-norm will bring more stable and efficient convergence than others.
At the same time, our strategy achieves 74.55\% top-1 accuracy on ImageNet, better than the inverse with 74.13\% and the N-M with 74.18\%.

\subsection{Performance Analysis}
\label{sec:performance}
\vspace{-0.1in}

\begin{figure*}[t]
    \begin{minipage}{0.247\linewidth}
        \centering
        \includegraphics[width=0.9\linewidth]{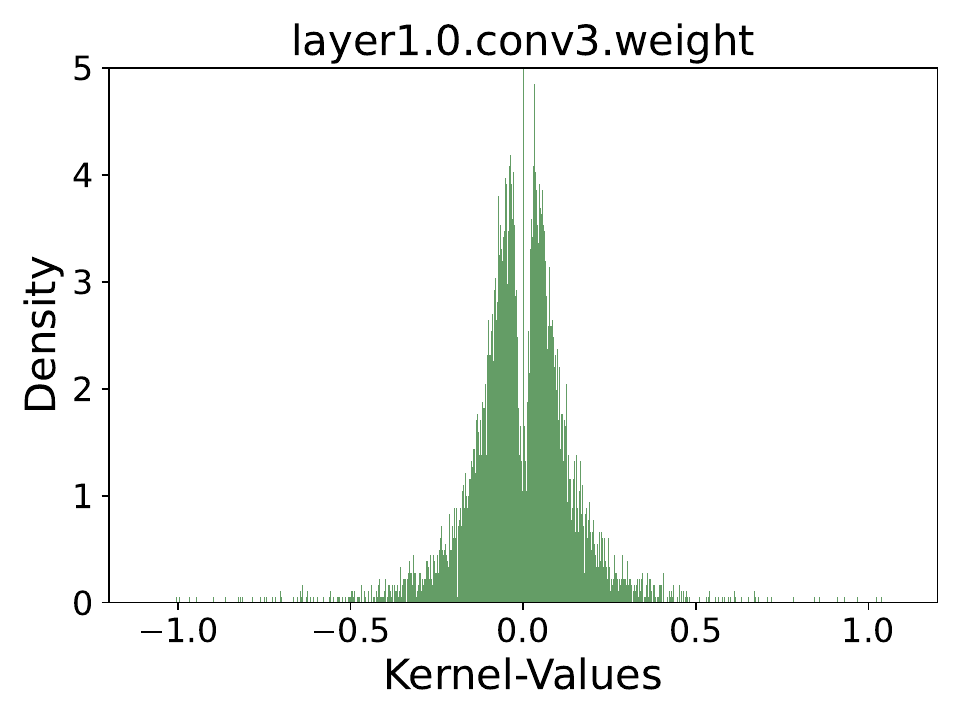}
    \end{minipage}
    \begin{minipage}{0.247\linewidth}
        \centering
        \includegraphics[width=0.9\linewidth]{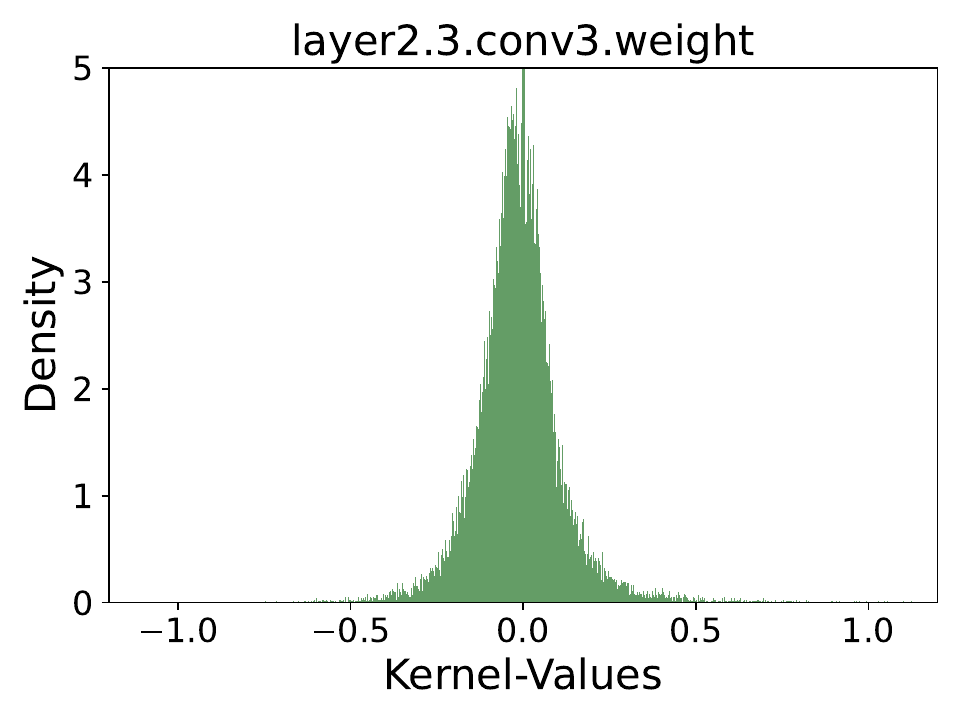}
    \end{minipage}
    \begin{minipage}{0.247\linewidth}
        \centering
        \includegraphics[width=0.9\linewidth]{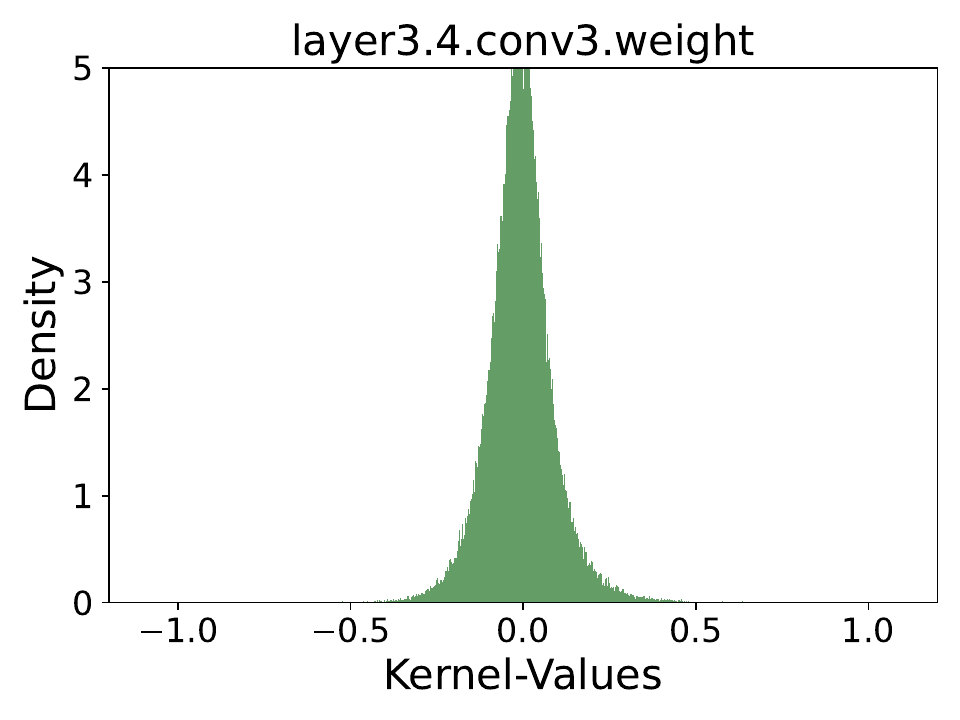}
    \end{minipage}
    \begin{minipage}{0.247\linewidth}
        \centering
        \includegraphics[width=0.9\linewidth]{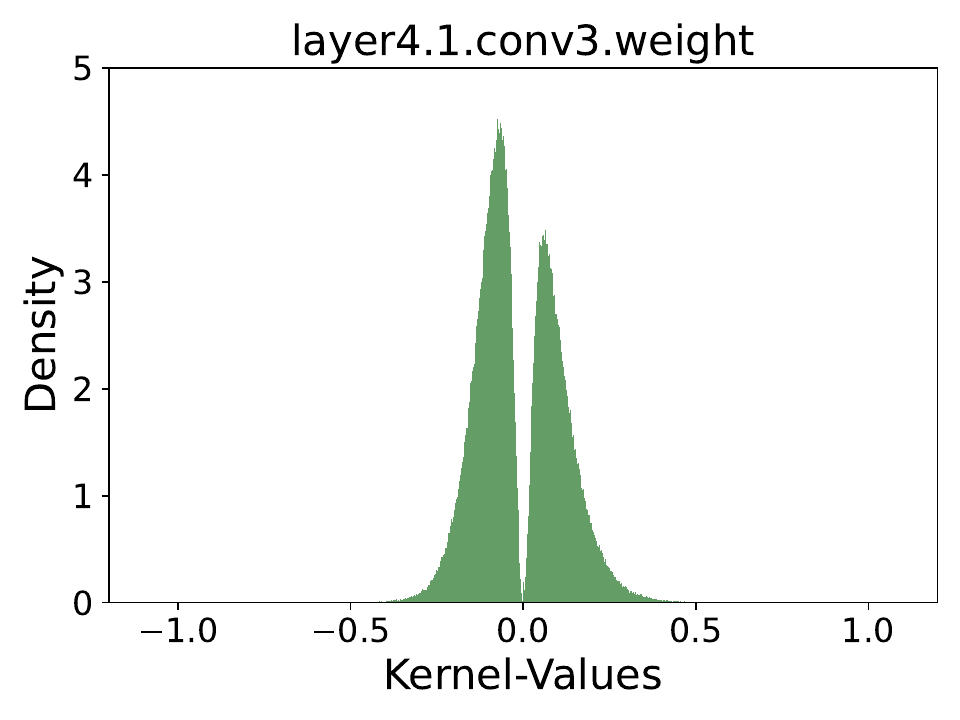}
    \end{minipage}
\vspace{-0.1in}
    \caption{Parameter distribution from SR-STE at inference.\label{fig:sr_ste_weight}}
\vspace{-0.1in}
\end{figure*}
\begin{figure*}[t]
    \begin{minipage}{0.247\linewidth}
        \centering
        \includegraphics[width=0.9\linewidth]{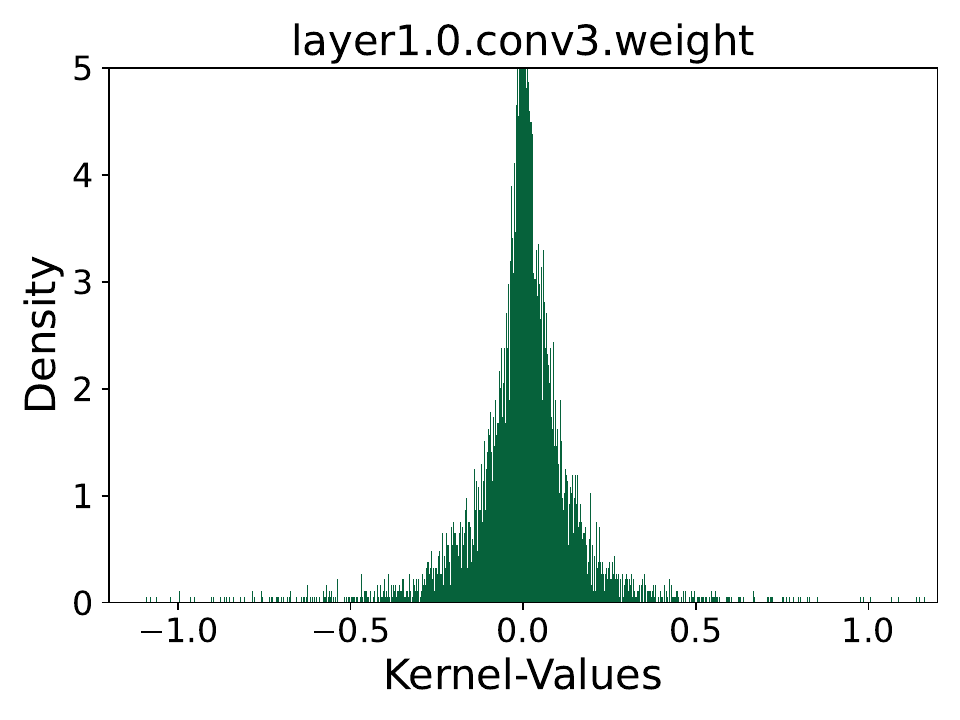}
    \end{minipage}
    \begin{minipage}{0.247\linewidth}
        \centering
        \includegraphics[width=0.9\linewidth]{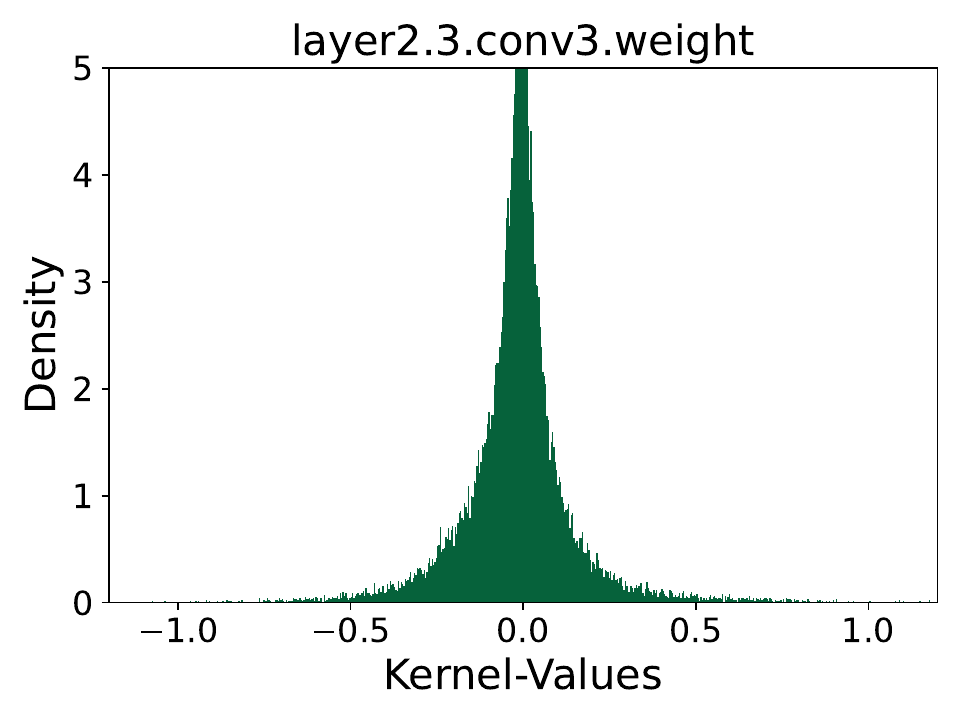}
    \end{minipage}
    \begin{minipage}{0.247\linewidth}
        \centering
        \includegraphics[width=0.9\linewidth]{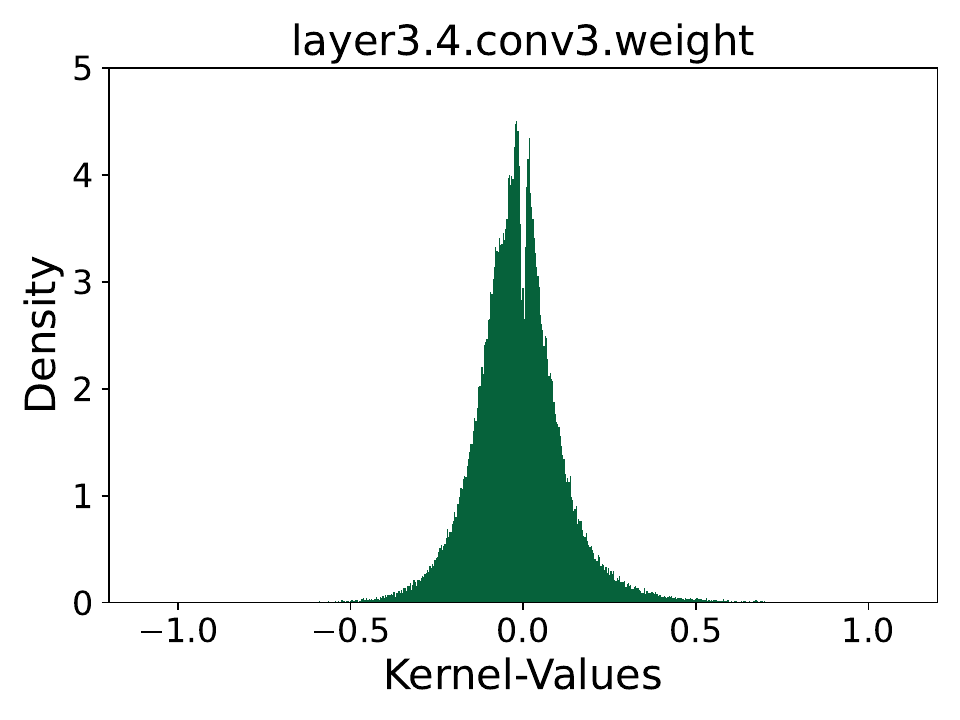}
    \end{minipage}
    \begin{minipage}{0.247\linewidth}
        \centering
        \includegraphics[width=0.9\linewidth]{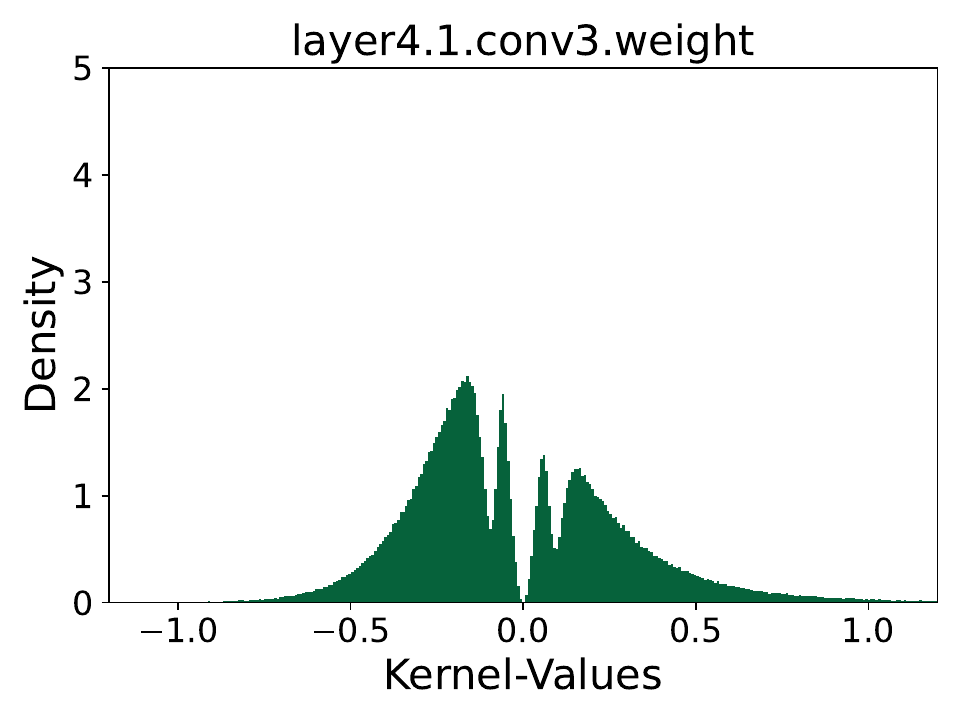}
    \end{minipage}
    \vspace{-0.1in}
    \caption{Parameter distribution from MaxQ at inference.\label{fig:maxq_weight}}
\vspace{-0.2in}
\end{figure*}

%

\begin{table}[h]
\scriptsize
\centering
\begin{tabular}{@{}lclcccc@{}}
\toprule
\multirow{2}{*}{Model}     & \multirow{2}{*}{N:M}  & \multirow{2}{*}{Method} & \multicolumn{2}{c}{Train speed} & \multirow{2}{*}{Top-1} & \multirow{2}{*}{\begin{tabular}[c]{@{}c@{}}FLOPs\\ (Train)\end{tabular}} \\ \cmidrule(lr){4-5}
                           &                       &                         & BS=128        & BS=256        &                        &                                                                          \\ \midrule
\multirow{10}{*}{ResNet50} & -                     & Dense                   & 798            & 884            & 77.3\%                   & 1$\times$(3.2e18)                                                         \\ \cmidrule(l){2-7} 
                           & \multirow{3}{*}{2:4}  & SR-STE                  & \bf{642}   & \bf{854}   & 77.0\%                   & 0.83$\times$                                                             \\
                           &                       & LBC                     & 373            & 487            & 77.2\%                   & \bf{0.72$\times$}                                                    \\
                           &                       & MaxQ                    & 507            & 732            & \bf{77.6\%}          & 0.91$\times$                                                             \\ \cmidrule(l){2-7} 
                           & \multirow{3}{*}{2:8}  & SR-STE                  & \bf{625}   & \bf{862}   & 76.2\%                   & 0.74$\times$                                                             \\
                           &                       & LBC                     & 382            & 512            & 76.5\%                   & \bf{0.53$\times$}                                                    \\
                           &                       & MaxQ                    & 514            & 743            & \bf{77.2\%}          & 0.86$\times$                                                             \\ \cmidrule(l){2-7} 
                           & \multirow{3}{*}{1:16} & SR-STE                  & \bf{628}   & \bf{852}   & 71.5\%                   & 0.69$\times$                                                             \\
                           &                       & LBC                     & 364            & 538            & 71.8\%                   & \bf{0.38$\times$}                                                    \\
                           &                       & MaxQ                    & 502            & 725            & \bf{74.6\%}          & 0.81$\times$                                                             \\ \bottomrule
\end{tabular}
\caption{Train speed~(NVIDIA RTX 3090, measured in samples/second/GPU), ImageNet top-1 accuracy and FLOPs~(train) with different N:M sparse pattern and methods.\label{tab:speed}}
\vspace{-0.2in}
\end{table}

\begin{table}[h]
\scriptsize
\centering
{
\begin{tabular}{@{}lccccc@{}}
\toprule
Model                     & N:M                  & Method & FP32    & INT8      & Acc Drop       \\ \midrule
\multirow{2}{*}{ResNet50} & \multirow{2}{*}{2:4} & MaxQ   & 77.6\%  & 77.1\%    & 0.5\%          \\
                          &                      & SR-STE & 77.1\%  & 76.6\%    & 0.5\%          \\ \bottomrule
\end{tabular}}
\caption{Post-Training Quantization~(PTQ) results on SR-STE and MaxQ for ResNet50 with 2:4 sparse pattern.\label{tab:quantization}}
\vspace{-0.1in}
\end{table}

\noindent
\textbf{Inference.}
As mentioned earlier, MaxQ can be considered a self-structured re-parameterization process during inference.
Thus, the soft masks and weights can be folded as constants and do not introduce any additional overhead during the inference.
Furthermore, as the soft masks also adhere to the N:M sparse pattern, 
MaxQ does not disrupt the sparse pattern and can still leverage its benefits, resulting in favorable latency on Ampere GPUs.

\noindent
\textbf{Training.}
The training efficiency of the algorithm is also an important aspect to consider.
While many algorithms have smaller FLOPs theoretically, they may not be highly parallelized or have a large amount of memory access.
Thus, there will be an increase but not a reduction in GPU days.
To investigate the efficiency of various algorithms for N:M sparsity, we present the train speed, top-1 accuracy, and training FLOPs for SR-STE, LBC, and MaxQ in \cref{tab:speed}.
For a fair comparison, their train speeds are tested with the same training script on the same machine with an NVIDIA RTX 3090 and automatic mixed precision.
SR-STE is the fastest.
MaxQ, although about 15\% slower than SR-STE, achieves the highest top-1 accuracy among the three methods.
The decrease in train speed primarily results from the multi-axis query, which incurs high memory access but involves minimal computational workload.
Furthermore, although LBC has the theoretically least training FLOPs, it is the slowest among the three methods. 
Compared to the train speed of dense network, LBC is almost 50\% slower for a batch size of 128 and 30\% slower for a batch size of 256.
In summary, MaxQ stands out as the most efficient algorithm, striking the best balance between training speed and accuracy.

\subsection{Quantization}
\vspace{-0.05in}

A structured re-parameterized network with simple PTQ may suffer from significant degradation in accuracy~\cite{chu2022make, ding2022re}, which is completely unusable.
To gain insights into the characteristics of MaxQ quantization, we first visualize the weight distribution of SR-STE and MaxQ in \cref{fig:sr_ste_weight} and \cref{fig:maxq_weight} respectively.
The weight distribution of SR-STE and MaxQ exhibits notable differences for the same model.
Specifically, in shallower layers, MaxQ has a sharper weight distribution.
While in deeper layers, it is smoother and has a broader dynamic range.
We conducted quantization experiments on SR-STE and MaxQ for comparison.
The results are presented in \cref{tab:quantization}.
To our surprise, our MaxQ demonstrates quantization-friendliness despite being perceived as a self-structured re-parameterized process.
For ResNet50, MaxQ with a 2:4 sparse pattern can achieve 0.5\% drop in the top-1 accuracy~(77.6\% $\rightarrow$ 77.1\%), when it is quantized to INT8 using a simple uniform PTQ method provided by the Ascend Tensor Compiler~(ATC).
Similar results are observed for SR-STE.
These experiments indicate that the weight distribution obtained by MaxQ are also applicable to quantize.
Furthermore, the performance of the INT8 model can still be improved by the more advanced quantization methods, such as non-uniform PTQ and Quantization-Aware Traning~(QAT).

\vspace{-0.05in}
\section{Conclusion}
\vspace{-0.05in}
N:M sparsity is a crucial method to reduce inference overhead and enable fast inference on NVIDIA Ampere GPUs.
In this paper, we propose a novel Multi-Axis Query method, MaxQ, to identify the critical weights and build a high-performance N:M sparsity network.
During the training, MaxQ employs a dynamic approach to generate soft N:M masks, which enhances the weights with more importance and ensures more effective updates for them.
During the runtime, soft N:M masks can be folded into the network as constants, which will not cause any distortion to the sparse pattern or bring additional computational costs.
Further, MaxQ follows a gradual sparse schedule by increasing the percentage of N:M weight blocks.
It progressively allows the network to heal from sparsity, avoiding severe information loss and achieving stable and efficient convergence.
Experiments on various vision tasks demonstrate that MaxQ can achieve notable performance gains over the state-of-the-art methods on multiple N:M sparse patterns and CNNs.

\noindent\textbf{Acknowledgements}
~This work is funded in part by the Key Research and Development Project of Zhejiang Province under Grant 2021C01035.

{
    \small
    \bibliographystyle{ieeenat_fullname}
    \bibliography{main}
}

\clearpage
\setcounter{page}{1}
\maketitlesupplementary

\section{MaxQ on ViT}

\subsection{Implentation Details}

\begin{table}[h]
\scriptsize
\centering
\setlength{\tabcolsep}{7mm}
\begin{tabular}{@{}l|c@{}}
\toprule
 & DeiT-Small  \\ \midrule
Stochastic depth survival prob & 0.90 \\
$t_i$ & 0 \\
$t_f$ & 225 \\
\midrule
Data augmentation & {rand-m9-mstd0.5-inc1} \\
Repeated Augmentation & {off} \\
Input resolution & {224} \\
Epochs & {300} \\
Batch size & {1024} \\
Warmup epochs & {20} \\
Hidden dropout & {0} \\
GeLU dropout & {0} \\
Attention dropout (if applicable) & {0} \\
Classification dropout & {0} \\
Random erasing prob & {0.25} \\
EMA decay & {0} \\
Cutmix $\alpha$ & {1.0} \\
Mixup $\alpha$ & {0.8} \\
Cutmix-Mixup switch prob & {0.5} \\
Label smoothing & {0.1} \\
Peak learning rate & {1e-3} \\
Learning rate decay & {cosine} \\
Optimizer & {AdamW} \\
Adam $\epsilon$ & {1e-6} \\
Adam $(\beta_1, \beta_2)$ & {(0.9, 0.999)} \\
Weight decay & {0.05} \\
Gradient clipping & {5.0} \\
\bottomrule
\end{tabular}
\caption{Hyperparameters for DeiT-Small on ImageNet-1K.}
\label{tab:hyper_deit}
\vspace{-0.1in}
\end{table}

\subsection{Results for ImageNet}

\begin{table}[ht]
\scriptsize
\centering
{
\begin{tabular}{@{}llccccc}
\toprule
Model                        & Method   & N:M         & Top-1           & Epochs & FLOPs     & Params    \\ \midrule
\multirow{4}{*}{DeiT-Small}  & Baseline & -           & 79.8\%          & 300    & 4.6G      & 22.1M      \\ \cmidrule(l){2-7} 
                             & SR-STE   & 2:4         & 75.7\%          & 300    & 2.5G      & 11.4M      \\
                             & LBC      & 2:4         & 78.0\%          & 300    & 2.5G      & 11.4M      \\ 
                             & \bf MaxQ & 2:4         & \bf 78.5\%      & 300    & 2.5G      & 11.4M      \\
\bottomrule            
\end{tabular}}
\caption{Results of the different N:M sparsity training methods for DeiT-Small on ImageNet.}
\label{tab:deit}
\vspace{-0.1in}
\end{table}

To further validate the effectiveness of MaxQ on Vision Transformer~(ViT), we conducted experiments with 2:4 sparsity on DeiT~\cite{touvron2021training}.
The hyperparameters and experiment results are shown in \cref{tab:hyper_deit} and \cref{tab:deit}.
MaxQ achieves 78.5\% top-1 accuracy at 2:4 sparse pattern while saving 45.6\% FLOPs and 48.5\% parameters.
Meanwhile, it exceeds SR-STE and LBC by 2.8\% and 0.5\% respectively.
It demonstrates that MaxQ is general and can enhance the performance of different types of deep neural networks.

\section{More Ablation Study}

\subsection{Incremental Schedulers}

\begin{table}[h]
\centering
\scriptsize
\begin{tabular}{@{}lccclc@{}}
\toprule
Model                     & N:M  & $t_i$ & $t_f$ & Scheduler                           & Top-1     \\ 
\midrule
\multirow{3}{*}{ResNet50} & 1:16 & 0     & 90    & cubic~(\cref{eq:incre_sparse})      & 74.6\%    \\
                          & 1:16 & 0     & 90    & linear(\cref{eq:linear})            & 74.5\%   \\
                          & 1:16 & 0     & 90    & cos~(\cref{eq:cos})                 & 74.3\%   \\ 
\bottomrule
\end{tabular}
\caption{Ablation study of different $t_i$ and $t_f$ in MaxQ.}
\label{tab:decay_schedule}
\end{table}

The ratio of N:M sparse blocks increases gradually with the training epoch.
We conduct several experiments for incremental schedulers to compare their effectiveness, including cubic~(default), linear~\cref{eq:linear} and cos~\cref{eq:cos} as follows:
\begin{equation}
    \delta_t=\min(1, \max(0, (t-t_i)/(t_f-t_i)))
\label{eq:linear}
\end{equation}
\begin{equation}
    \delta_t=
    \begin{cases}
        0, & t\leq t_i \\
        1-\frac{1}{2}\left(1+\cos(\frac{t-t_i}{t_f-t_i}\pi)\right), & t_i < t\leq t_f \\
        1, & t_f<t
    \end{cases}
\label{eq:cos}
\end{equation}

\begin{figure}[t]
\centering
\includegraphics[width=0.85\linewidth]{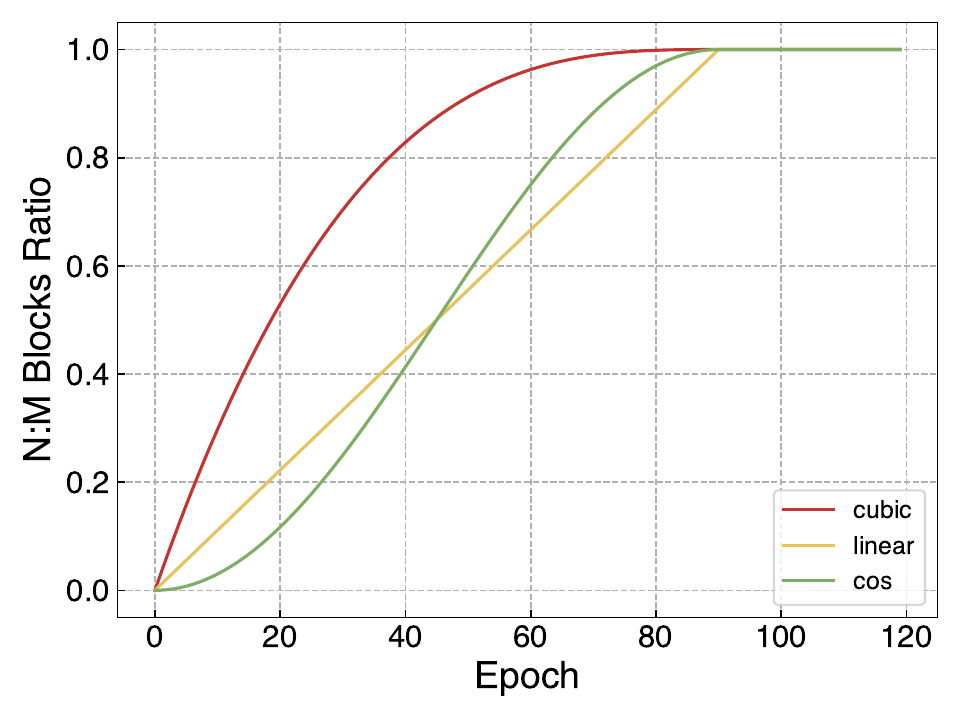}
\caption{Visualization for N:M blocks ratio with different incremental schedulers.}
\label{fig:schedule}
\end{figure}

As shown in \cref{tab:decay_schedule}, cubic scheduler~(default) performs better than the other schemes by 0.1\% and 0.3\% top-1 accuracy.
We draw the N:M blocks ratio change for these three schemes in \cref{fig:schedule}.
It suggests that rapidly increasing the ratio of N:M sparse blocks at the beginning of training will facilitate the model's convergence and achieves better performance.
%

\section{Optimization}

For back propagation, MaxQ follows the SR-STE
\begin{eqnarray}
\begin{split}
    \mathbf{m}^l_{t+1} & = \mathbf{m}^l_t-
    \gamma_t[g\left(\mathbf{s}^l_t\odot \mathbf{m}_t^l\right)  \\
    & +\sigma\left(1-\mathop{clip}\left(\mathbf{s}_t^l,0,1\right)\right)\odot \mathbf{m}_t^l]
\end{split}
\end{eqnarray}
where $\gamma_t$ is the learning rate for the t-step, 
$\sigma$ is the denotes the relative weight for the sparse-refined term and we set them to 2$\times$weight\_decay, 
$g$ is the gradient function and we estimate the gradient for mask operator by straight-through estimator~(STE). 
Meanwhile, we clip the $\mathbf{s}^l$ to $[0,1]$ for avoiding negative values.


\end{document}